\documentclass{article}

\usepackage{PRIMEarxiv}

\usepackage[utf8]{inputenc} 
\usepackage[T1]{fontenc}    
\usepackage{hyperref}       
\usepackage{url}            
\usepackage{booktabs}       
\usepackage{amsfonts}       
\usepackage{nicefrac}       
\usepackage{microtype}      
\usepackage{lipsum}
\usepackage{fancyhdr}       
\usepackage{graphicx}       
\graphicspath{{media/}}     

\newcommand{\myClassifier}{EVVER-Net}
\newcommand{\myDataset}{CREDULE}

\usepackage{graphicx}
\usepackage{booktabs}
\usepackage{tabularx}
\usepackage{caption}

\usepackage{amsmath}

\title{Credible, Unreliable or Leaked?: Evidence Verification for Enhanced Automated Fact-checking}

\usepackage{authblk}

\author[1]{\small Zacharias Chrysidis}
\author[1, 2]{\small Stefanos-Iordanis Papadopoulos}
\author[2]{\small Symeon Papadopoulos}
\author[1]{\small Panagiotis C. Petrantonakis}

\affil[1]{\footnotesize Department of Electrical \& Computer Engineering, Aristotle University of Thessaloniki.}
\affil[2]{\footnotesize Information Technology Institute, Centre for Research \& Technology, Hellas.}

\affil[ ]{\textit{zachoschrissidis@gmail.com},
\{\textit{stefpapad,papadop\}@iti.gr, \textit{ppetrant@ece.auth.gr}
}}

\begin{document}
\maketitle

\begin{abstract}
Automated fact-checking (AFC) is garnering increasing attention by researchers aiming to help fact-checkers combat the increasing spread of misinformation online. 
While many existing AFC methods incorporate external information from the Web to help examine the veracity of claims, they often overlook the importance of verifying the source and quality of collected ``evidence''. 
One overlooked challenge involves the reliance on ``leaked evidence'', information gathered directly from fact-checking websites and used to train AFC systems, resulting in an unrealistic setting for early misinformation detection.
Similarly, the inclusion of information from unreliable sources can undermine the effectiveness of AFC systems.
To address these challenges, we present a comprehensive approach to evidence verification and filtering. 
We create the ``CREDible, Unreliable or LEaked'' (\myDataset) dataset, which consists of 91,632 articles classified as Credible, Unreliable and Fact-checked (Leaked).
Additionally, we introduce the EVidence VERification Network (\myClassifier), trained on \myDataset\ to detect leaked and unreliable evidence in both short and long texts. \myClassifier\ can be used to filter evidence collected from the Web, thus enhancing the robustness of end-to-end AFC systems. 
We experiment with various language models and show that \myClassifier\ can demonstrate impressive performance of up to 91.5\% and 94.4\% accuracy, while leveraging domain credibility scores along with short or long texts, respectively.
Finally, we assess the evidence provided by widely-used fact-checking datasets including LIAR-PLUS, MOCHEG, FACTIFY, NewsCLIPpings+ and VERITE, some of which exhibit concerning rates of leaked and unreliable evidence. 
\end{abstract}


\section{Introduction}
\label{Sec: Intro}
Misinformation has become an increasingly prevalent issue today, causing negative impacts to individuals and society \cite{duffy2020too}. With the rapid spread of online platforms and social media, fake or misleading information has become alarmingly widespread, posing significant challenges to informed decision-making and societal trust \cite{olan2022fake}.
In the battle against misinformation, many fact-checking platforms such as Snopes\footnote{Snopes: \url{https://www.snopes.com/}},  PolitiFact\footnote{Politifact: \url{https://www.politifact.com/}} and Reuters \footnote{Reuters: \url{https://www.reuters.com/fact-check/}} have emerged, where journalists manually review a plethora of claims sourced from news articles and social media. Nonetheless, manual fact-checking is time-consuming and can not always keep pace with the rate at which misinformation spreads.
Recently, researchers in natural language processing \cite{oshikawa2018survey}, computer vision \cite{zhao2021multi} and multimodal learning \cite{alam2021survey} have begun exploring Automated Fact-Checking (AFC).
AFC involves tools and systems that 
help professional fact-checkers to combat misinformation more efficiently by automating pivotal aspects of fact-checking including claim detection, evidence retrieval and claim verification \cite{guo2022survey}.

\begin{figure*}
    \includegraphics[width=\textwidth]{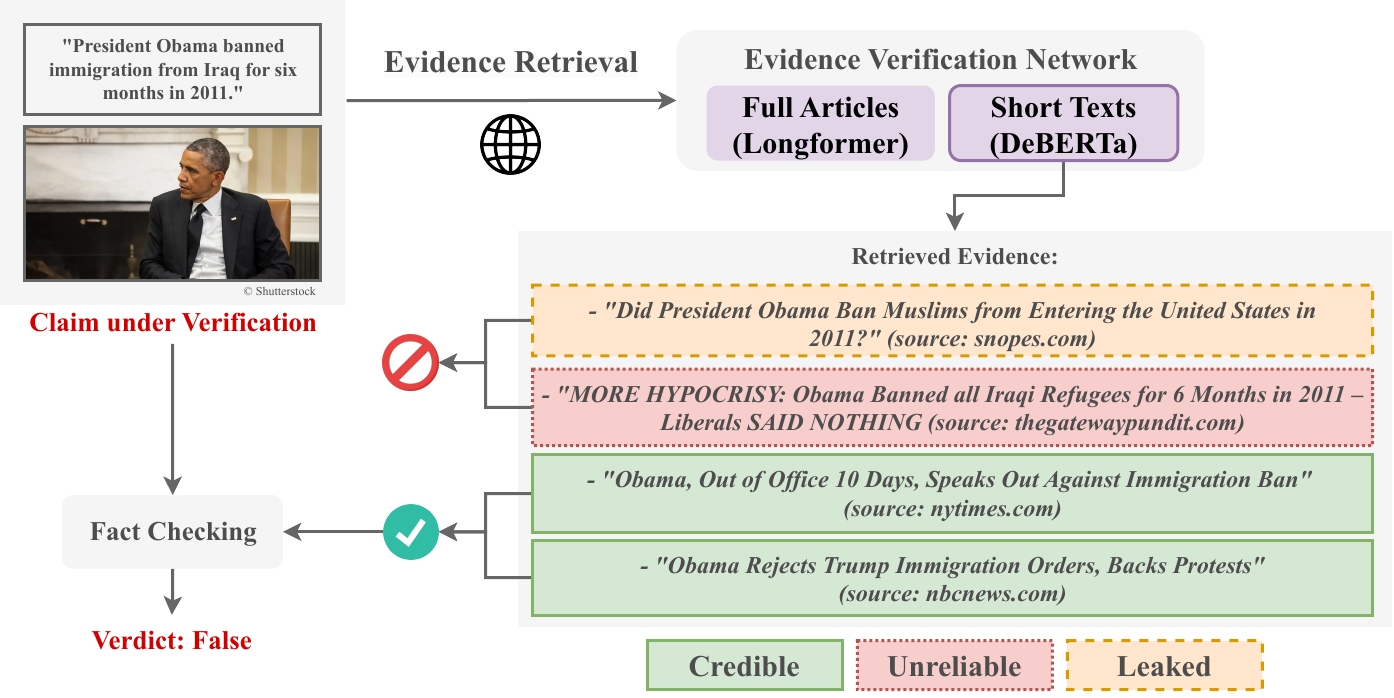}
    \caption{Pipeline of automated fact-checking leveraging the proposed Evidence Verification Network.}
    \label{fig:teaser}
\end{figure*}

To further improve AFC, some systems leverage external evidence extracted from the Web using search engines. By tapping the potential of the entire Web as a knowledge source to help support or refute a claim, these models enhance their ability to verify news pieces more accurately.
However, a prevalent issue in external knowledge retrieval from the Web is the lack of evidence filtering mechanisms. 
Inadequate or rudimentary filtering results in the inclusion of irrelevant or unreliable information, potentially compromising the accuracy of fact-checking systems.
Furthermore, Glockner et al. \cite{glockner-etal-2022-missing}
define \textit{``two requirements that the evidence in datasets must fulfill for realistic fact-checking: It must be (1) sufficient to refute the claim and (2) not leaked from existing fact-checking articles''}.
Otherwise
the AFC model would learn to rely on previously fact-checked information when trying to detect new emerging misinformation, where fact-checks are not yet available. 
The problem of ``leaked evidence'' is quite under-researched yet crucial for realistic and effective fact-checking.

Motivated by these observations, we propose a new evidence verification and filtering approach to address the issue of leaked and unreliable evidence in AFC. Firstly, we construct the ``CREDible, Unreliable or LEaked'' (\myDataset) dataset, by modifying, merging, and extending MultiFC \cite{augenstein2019multifc}, Politifact \cite{politifact_kaggle_dataset}, PUBHEALTH \cite{kotonya2020explainable}, NELA-GT \cite{horne2018sampling,norregaard2019nela,gruppi2020nelagt2019,gruppi2021nela,gruppi2020nelagt2021,gruppi2023nelagt2022}, Fake News Corpus \cite{fakenewscorpus}, and  Getting Real About Fake News \cite{megan_risdal_2016}. These established datasets contain short texts (titles) as well as the long texts (full articles) of the news articles. 
We extract the article bodies, where they are not given, and other meta-data to better balance the classes. 
The final \myDataset\ dataset consists of 91,632 pieces, equally distributed in three classes: ``Credible'', ``Unreliable'' and ``Fact-checked'' (or Leaked). 

The goal is to develop a model capable of detecting the information that a model retrieves from the Web so as to avoid leakage and unreliable sources, as seen in Figure \ref{fig:teaser}. To this end, we also propose \myClassifier\, a neural network that detects leaked (fact-checked) and unreliable evidence pieces during the evidence retrieval process and only allows credible information to pass to the AFC model. 
We experiment with various pre-trained Transformer-based encoders for both short texts, such as DeBERTa \cite{he2020deberta}, CLIP \cite{radford2021learning-clip}, T5 \cite{raffel2020exploring-t5}, and long texts, Long T5 \cite{guo2021longt5} and Longformer \cite{beltagy2020longformer}, as well as baseline methods like Count Vectorizer and TF-IDF.
Additionally, we integrate domain credibility scores from the Media Bias/Fact Check (MBFC) website \footnote{\url{https://mediabiasfactcheck.com/}} to enhance classification accuracy.

To show the efficacy and usefulness of the classifier, we examine the collected evidence in widely used AFC datasets such as the LIAR-PLUS \cite{alhindi-etal-2018-evidence}, FACTIFY \cite{mishra2022factify}, MOCHEG \cite{mocheg}, VERITE \cite{papadopoulos2024verite} 
datasets and the evidence collected by Abdelnabi et al. \cite{abdelnabi2022open} for the  NewsCLIPpings dataset; referred to as NewsCLIPpings+ for simplicity. 
Our analysis shows that the collected evidence often contain information leaked from fact-checking articles or provide unreliable information.

The contributions of our work can be summarized as follows:
\begin{itemize}

\item We propose a novel approach to detecting and filtering out leaked evidence and unreliable information in AFC systems.

\item We construct \myDataset, a large-scale, balanced and diverse dataset comprising ``Credible'', ``Unreliable'' and ``Fact-checked'' news articles\footnote{We release our code at: \url{https://github.com/mever-team/credule-dataset}}.

\item We introduce \myClassifier\, which demonstrates impressive performance of up to 91.5\% and 94.4\% accuracy on \myDataset\, while leveraging domain credibility scores along with short or long texts, respectively. 

\item We use \myClassifier\ to examine the evidence of widely used fact-checking datasets, where we identify concerning rates of leaked and unreliable evidence.

\end{itemize}

\section{Related Work}

\subsection{Fact-checking Datasets}

A plethora of datasets have been curated to facilitate fact-checking tasks and train robust misinformation detection models. 
One widely used text-based dataset is FEVER \cite{thorne-etal-2018-fever}.
It consists of 185,445 pieces generated by human annotators extracting claims from Wikipedia and mutating them in various ways, some of which alter their meaning.
Many datasets also contain claims extracted from fact-checking websites. For instance, Alhindi et al. \cite{alhindi-etal-2018-evidence} created the LIAR-PLUS dataset, comprising 12,836 statements taken from Politifact and labeled by humans for truthfulness. The authors also automatically extracted justifications provided in the associated fact-checking articles.
Others include FakeNewsNet \cite{shu2020fakenewsnet}, MultiFC \cite{augenstein2019multifc}, Politifact Fact Check \cite{politifact_kaggle_dataset} or WatClaimCheck \cite{khan2022watclaimcheck} and some specialize in various domains such as politics (ClaimBuster \cite{hassan2017claimbuster} or Truth of Varying Shades \cite{rashkin2017truth}) or health (PUBHEALTH \cite{kotonya2020explainable}).
Additionally, researchers have created datasets containing both credible and fake news pieces. For example, the NELA-GT Datasets (2017-2022) \cite{horne2018sampling,norregaard2019nela,gruppi2020nelagt2019,gruppi2021nela,gruppi2020nelagt2021,gruppi2023nelagt2022} are large corpora containing articles from both credible and non-credible news outlets. FakeNewsCorpus \cite{fakenewscorpus} contains both credible and fake news scraped from a curated list of 1001 non-credible domains. Conversely, the Getting Real about Fake News Dataset \cite{megan_risdal_2016} extracted articles from 244 websites tagged as ``bullshit'' by the BS Detector Chrome Extension.

Multimodal datasets play a vital role in AFC by incorporating diverse types of information. 
They offer a more comprehensive representation of real-world claims, enabling fact-checking models to consider a broader range of evidence sources.
MOCHEG \cite{mocheg} is a dataset consisting of 21,184 textual claims from Politifact and Snopes that also provides image and textual evidence collected from fact-checking articles.
Focusing on social media content, Boididou et al. \cite{boididou2018verifying} built a dataset for the MediaEval 2016 Verifying Multimedia Use (VMU) challenge that comprises tweets and images.
Another notable multimodal dataset is FACTIFY \cite{mishra2022factify}, containing 50,000 claims accompanied by 100,000 images. Collected from reliable US and Indian sources, as well as reputable fact-check websites, FACTIFY provides a diverse range of real-world data for fact-checking purposes.
Researchers have also been experimenting with synthetically created multimodal misinformation. Aneja et al. \cite{aneja2021cosmos} curated COSMOS, a dataset comprising 200K images with 450K textual captions from various news websites (credible and fact-check), blogs, and social media posts and randomly sampled negative ``de-contextualized'' samples. 
Similarly, the NewsCLIPpings dataset \cite{luo2021newsclippings} comprises both pristine and convincing falsified (‘out-of-context’) image-caption pairs, providing examples of how misinformation can be spread through visual content. 
But instead of relying on ``naive'' random negative samples, the authors leverage CLIP \cite{radford2021learning-clip} as well as Person and Scene Matching models to create ``hard'' negative samples. 
Built on the VisualNews corpus \cite{liu2020visual}, NewsCLIPpings contains examples that misrepresent the context, place, or people in the image.
Finally, the VERITE dataset was recently developed as an evaluation benchmark for multimodal misinformation detection and accounts for unimodal biases \cite{papadopoulos2024verite}.

\subsection{Evidence Collection}

In the pursuit of enhancing the effectiveness of AFC, researchers have recognized the value of gathering and utilizing external information from the Web. Models often leverage popular search engines to access a vast repository of information that can supplement existing datasets.
Popat et al. \cite{popat2016credibility} utilized claims from Snopes and information about hoaxes and fictitious persons from Wikipedia to conduct their experiments, employing these claims and hoaxes 
as queries to the Google search engine. 
In subsequent work, the authors attempted to rank  results based on the credibility of their sources \cite{popat2018declare}.
Samarinas et al. \cite{factualnli} extended the FEVER dataset \cite{thorne-etal-2018-fever} to create the Factual-NLI+ dataset, incorporating synthetic examples and noise passages from web search results. Retrieving the top 30  results from the Bing Search engine for each claim in the FEVER dataset, they retained results with the highest BM25 score.
Similarly, Abdelnabi et al. \cite{abdelnabi2022open} collected external information from the Web to verify image-caption claims in the NewsCLIPpings \cite{luo2021newsclippings} dataset. Employing an inverse search mode via the Google Vision APIs, they retrieved textual evidence such as text snippets and image captions and then utilized the caption as textual queries to search for images using the Google Custom Search API.
More recently, a similar approach was adopted to augment the VERITE evaluation benchmark \cite{papadopoulos2024verite} with external information from the Web \cite{papadopoulos2023red}. 

\subsection{Addressing Leaked Evidence}

While leveraging external evidence from the Web holds promise for enhancing the accuracy of AFC systems, it also presents challenges, particularly concerning the presence of leaked evidence. The criteria outlined by \cite{glockner-etal-2022-missing} stress the importance of ensuring that evidence used is not leaked from existing fact-checking articles.
However, relying on search engines to retrieve external evidence often results in leaked information being included unintentionally. Even after excluding search results that point to the claim’s fact-checking article, leaked evidence persists. This can occur when different organizations verify the same claims or disseminate fact-checkers' verifications. 
Khan et al. \cite{khan2022watclaimcheck} also highlight this issue, noting that `premise' articles may indirectly leak the veracity label.
Glockner et al. \cite{glockner-etal-2022-missing} express doubts on whether automated approaches can realistically refute harmful real-world misinformation as many of existing approaches fail to overcome the information leakage problem. This underscores the need for robust mechanisms to filter out leaked evidence and maintain the credibility of AFC processes.

\subsection{Evidence Filtering}

Filtering and verifying external evidence pose significant challenges for AFC systems. Many existing models lack robust filtering mechanisms or rely on rudimentary approaches, potentially resulting in the inclusion of irrelevant, untrustworthy, or leaked evidence. Addressing this challenge requires the development of advanced filtering techniques capable of discerning reliable sources and accurate information.
For example, Abdelnabi et al. \cite{abdelnabi2022open} implemented a filtering method that discards evidence items matching the query and originating from the same website, utilizing techniques such as removing punctuation and converting captions to lowercase for textual evidence and employing perceptual hashing for images.
Karadzhov et al. \cite{karadzhov2017fully}, focusing on the trustworthiness of the source, disregarded evidence from domains considered unreliable based on manual checks of the most frequent domains in search results. However, this approach may not effectively identify unreliable sources that appear less frequently.
Popat et al. \cite{popat2016credibility} adopted an approach to assess the reliability of Web sources by determining the AlexaRank and PageRank of each source. AlexaRank measures website popularity based on its unique visitors and page views, while PageRank assesses website importance by considering the number and quality of links to and from the website. 
While such approaches provide valuable insights into the authority and popularity of Web sources, they may not accurately reflect their credibility from a fact-checking standpoint.
In their recent study, Schlichtkrull et al. \cite{schlichtkrull2024averitec}, employed a custom Google search tool to mitigate temporal leaks, ensuring that only documents published before the claim date were retrieved. 
However, this approach can introduce noise, as the dates provided by Google Search are not consistently accurate.

\section{METHODOLOGY}

\subsection{Problem Formulation}
 
The AFC process typically unfolds through three sequential stages: claim detection, evidence retrieval, and claim verification, as outlined by Guo et al. \cite{guo2022survey}. 
However, a fundamental challenge arises in the evidence retrieval stage, as not all available information can be considered trustworthy. Guo et al. \cite{guo2022survey} underscore this issue, highlighting that reliance on single authoritative sources may overlook contradicting or untrustworthy evidence. Additionally, Glockner et al. \cite{glockner-etal-2022-missing} emphasize the problem of information leakage, where existing AFC models incorporate leaked evidence, compromising the integrity and realism of the fact-checking process.

In response to these challenges, we construct \myDataset, a large scale dataset containing news articles from various sources. These articles are classified in three classes: ``Credible'', ``Unreliable'' and ``Fact-checked''.
We also create the EVidence VERification Network (\myClassifier) and train it on \myDataset, with the aim to detect leaked and unreliable evidence and only allow credible information to be examined by an AFC system. 
Specifically, the \myClassifier\ $\mathcal{G(\cdot)}$ aims to classify each textual evidence snippet $e_i$ as fact-checked(0), credible(1), or unreliable(2), denoted by $\mathcal{G}(e_i)$. 
Let $C$ denote the claim under scrutiny, 
$E_T  = \{ e_1, e_2, \ldots, e_N \}$
represents the set of $N$ textual evidence pieces and $E_I$ symbolizes the collection of image evidence related to $C$ . 
After applying the classifier $\mathcal{G(\cdot)}$, we obtain the filtered set of textual evidence pieces $E_T'$, denoted as:
\begin{equation}
\label{eqn:e_t}
E_T' = \{ e_i \mid \forall e_i \in E_T, \mathcal{G}(e_i) = 1 \}
\end{equation}

Subsequently, the veracity prediction mechanism $\mathcal{F(\cdot)}$ can be applied to the claim $C$ with the filtered text evidence $E_T'$ and the image evidence $E_I$:
\begin{equation}
\label{eqn:y}
y = \mathcal{F}(\mathcal{C}, E_T', E_I)
\end{equation}
where $y$ represents the predicted veracity label.
$\mathcal{G}(\cdot)$ serves as a critical component in mitigating the impact of trustworthiness issues and information leakage in AFC systems.

\subsection{Constructing the \myDataset\ dataset}

\myDataset\ comprises three distinct classes: Fact-Checked, Credible, and Unreliable, tailored for the classification of news articles to facilitate evidence filtering during the fact-checking process. Ensuring a balanced representation, each class encompasses a similar number of articles, consistent distributions in publication year, title length, and thematic content. Maintaining an even distribution of articles across years ensures comprehensive coverage of events over time. Entries in our dataset are presented in both article title and full-text formats whenever available.

To construct \myDataset, we merge information from six distinct sources: MultiFC \cite{augenstein2019multifc}, Politifact Fact-Check \cite{politifact_kaggle_dataset}, PUBHEALTH \cite{kotonya2020explainable}, NELA-GT \cite{horne2018sampling,norregaard2019nela,gruppi2020nelagt2019,gruppi2021nela,gruppi2020nelagt2021,gruppi2023nelagt2022}, Fake News Corpus \cite{fakenewscorpus}, and  Getting Real About Fake News \cite{megan_risdal_2016} datasets. These diverse datasets provide a rich and comprehensive foundation for our dataset creation process.

\subsubsection{Fact-checked class}

In order to construct the fact-checked class, we leverage the following dataset: 
1) \textbf{MultiFC:} encompassing 36,534 claims sourced from 24 Fact-checked domains, spanning until 2019. Each claim is accompanied by its respective veracity label, publication date, article title, URL, and additional metadata. 
2) \textbf{PUBHEALTH:} comprising 11,832 claims within the health domain, sourced from various fact-checking websites, this dataset offers valuable insights. Alongside each claim, the dataset provides the article's URL, publication date, and the main text body.
3) \textbf{Politifact:} We incorporate 21,152 statements from Politifact, spanning the years 2008 to 2022, sourced from the Politifact Fact Check dataset on Kaggle. Each entry includes URLs, truthfulness labels, and additional metadata. To ensure data consistency, we extract article titles from the URLs, omitting the term ``Politifact'' to mitigate biases. 

From our collection of fact-checking articles, we retain only those published between 2016 and 2022. We also filter out duplicate and empty entries, resulting in 30,209 articles. Additionally, we employ a pre-trained DistilBert model, trained on the News Category Dataset\footnote{HuggingFace Model: \url{https://huggingface.co/Yueh-Huan/news-category-classification-distilbert}}, to extract article topics based on their titles. This process yields a comprehensive set of 42 categories, which we further consolidate into 12 distinct groups for clarity.
Furthermore, we extract the full text-body of articles where unavailable. 
First, we utilize the Beautiful Soup package to obtain the full article texts from the Politifact Fact Check dataset. 
Similarly, we develop domain-specific scripts to extract articles from the MultiFC dataset,  excluding those already present in the Politifact dataset. 
The PUBHEALTH dataset provides full-text articles, obviating the need for extraction.

\begin{figure}
  \centering
  \includegraphics[width=0.7\textwidth]{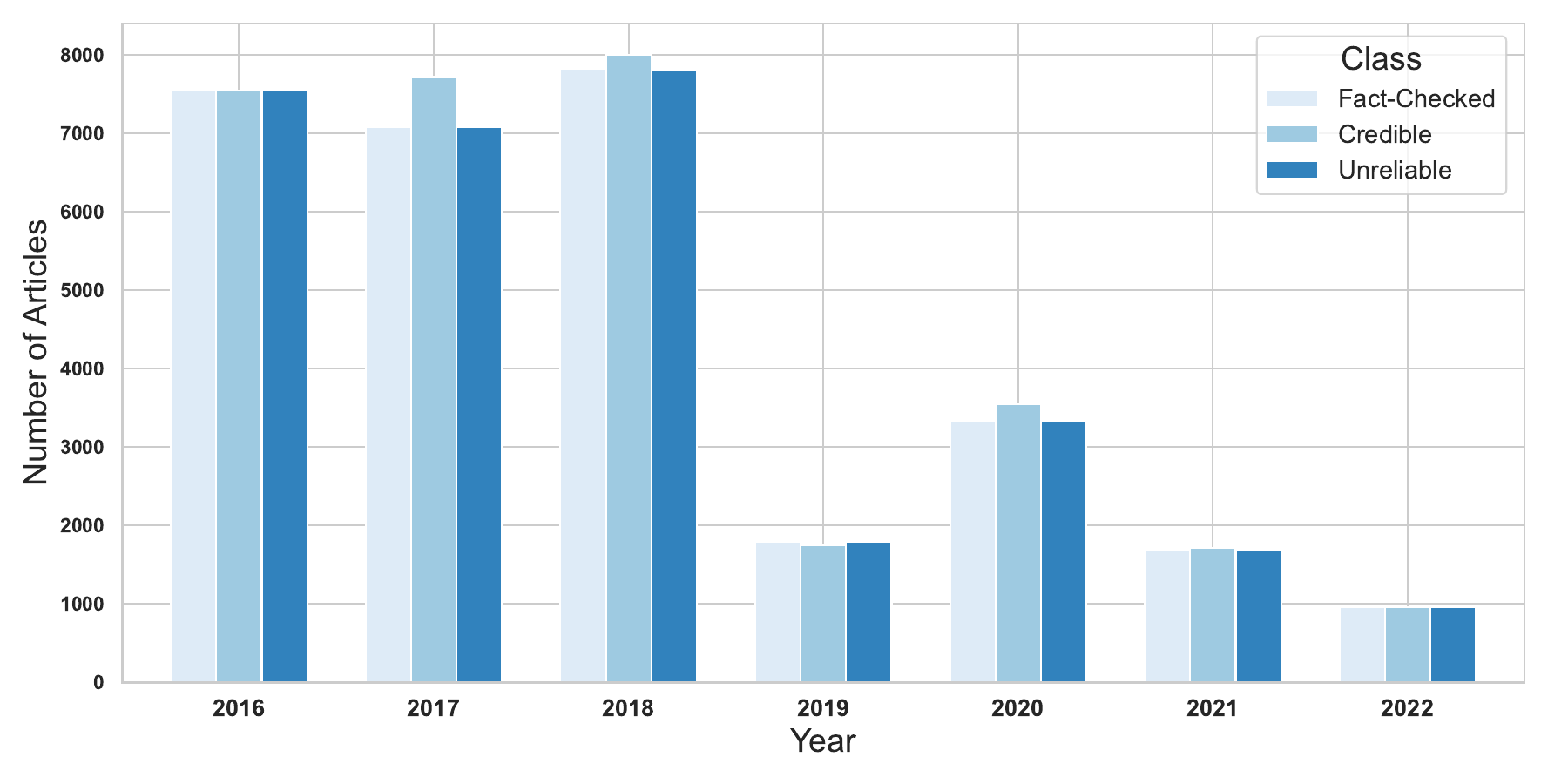}
  \caption{Number of articles per year in \myDataset.}
  \label{fig:articles_per_year_topic.png}
\end{figure}

\subsubsection{Credible and Unreliable Classes}
To construct the Credible and Unreliable classes, we leverage the following datasets: 
1) \textbf{Fake News Corpus:} It comprises millions of news articles, including both credible and non-credible sources. We specifically utilize the credible subset, extracted from reputable news outlets. To ensure data integrity, we extract article titles and dates directly from URLs, addressing issues with inaccuracies and duplicates. Additionally, we remove domain names from titles and employ the topic-extraction model to categorize the articles.
2) \textbf{NELA-GT Datasets:} Spanning from 2017 to 2022, these contain articles sourced from various domains, classified based on their credibility and bias. We extract pertinent information such as article titles, dates, domains, and topics using the topic-extraction model. These datasets are instrumental in constructing both the credible and non-credible classes.
3) \textbf{Getting Real about Fake News Dataset:} The Getting Real about Fake News dataset (GRAFN), sourced from Kaggle, features non-credible articles from 2016, scraped from 244 websites labeled as ``bullshit'' by the BS Detector Chrome Extension. We filter the dataset to include only English articles categorized as `bs' (bullshit), `conspiracy', `satire', `junksci', and `fake'. Obtaining article metadata, including titles, dates, domains, and main text bodies, enables us to further categorize articles based on their topics.

To ensure class balance across different years (2016-2022) and topics, we supplement articles from the Fake News Corpus and, if necessary, the NELA-GT datasets for the Credible class. 
We ensure an equal distribution of articles across various topics to align with the Fact-checked class, extracting the same number of articles per topic.
Furthermore, for the Unreliable class, we incorporate articles from both NELA-GT and Getting Real About Fake News datasets.
The distribution of articles across different years and classes, can be seen in
Figure \ref{fig:articles_per_year_topic.png}. 
\myDataset\ is balanced across classes in terms of articles per year and in terms of the length of titles within each class.
We make use of the full text articles provided by NELA-GT and Getting Real About Fake News datasets. For the Fake News Corpus, we utilize the Beautiful Soup library and develop custom scripts to extract the content of full articles.

\begin{table}
\centering
    \caption{\myDataset\ dataset statistics, constructed by combining and filtering various datasets.}
  \label{tab:Dataset-Statistics} 
  \begin{tabular}{cccc}
    \toprule
    Dataset&Class&Used Entries&Full-Texts\\
    \midrule
     MultiFC & Fact-checked & 16,057 & 70.5\%\\
     PUBHEALTH & Fact-checked & 3,683 & All\\
     Politifact & Fact-checked & 10,469 & 99.7\%\\
     Fake News Corpus & Credible & 22,245 & 91.4\%\\
     NELA-GT & Credible+Unreliable & 31,636 & All \\  
     GRAFN & Unreliable & 7,542 & All\\ 
  \bottomrule
\end{tabular}
\end{table}

\subsubsection{Domain Credibility Scores (DCS)}\label{subsec:mbfc}

We augment \myDataset\ by integrating external domain credibility scores (DCS)
from the Media Bias/Fact Check (MBFC) website. 
MBFC is an independent platform dedicated to combating media bias and misinformation and employs a rigorous evaluation combining objective metrics and subjective analysis to rate media sources based on factors such as bias, factual accuracy, and overall credibility.
Bias assessments range from least biased to extreme bias on a scale of 0 to 10, while factuality scores vary from very high to very low based on fact-checking frequency and inclusion of critical information.
Categorized into three tiers—high, medium, and low credibility—the final MBFC rating identifies highly credible sources with a score of 6 or above, medium credibility for scores ranging from 3 to 5, and low credibility for scores of 0 to 2 or sources rated as questionable, conspiracy, or pseudoscience.
For each domain in \myDataset, we obtain DCS including `Bias Rating', `Factual Reporting' and `MBFC Credibility Rating'. 
This process involves querying the MBFC website for each domain, accessing the corresponding page, and extracting the relevant scores. 
We are able to extract domain credibility scores for 92.2\% of the articles in \myDataset.

\subsubsection{\myDataset\ Final Statistics}

The \myDataset\ comprises a total of 91,632 articles, spanning the years 2016 to 2022 and classified into three distinct categories: Fact-checked (30,209), Credible (31,230), and Unreliable (30,193). Each article entry includes its title, publication date, URL, assigned topic, and classification label. 
Moreover, 92.7\% of the articles in the dataset are accompanied by their full-text content. Notably, the dataset exhibits balanced distribution across all classes in terms of articles per year, topics, and title lengths. The statistics of the dataset are summarized in Table \ref{tab:Dataset-Statistics}.

\subsection{Evidence Verification Network}

After constructing \myDataset, we develop an EVidence VERification Network (\myClassifier) which can be used to discern between credible, unreliable and leaked evidence. 
\myClassifier\ is a neural network classifier, which can be expressed as follows:
\begin{equation}
\label{eqn:model}
\hat{y_i}=Softmax(\textbf{W}_1\cdot \text{GELU}(\textbf{W}_0\cdot [T(e_i)[< EOS >];s_i])))
\end{equation}
where $T(\cdot)$ stands for a Transformer backbone encoder, $[<EOS>]$ for the position of the end-of-sentence (EOS) or classification  token (CLS), depending on the encoder, of $T(\cdot)$, $s_i\in\mathbb{R}^{1}$ is the ``domain credibility score'' of $e_i$, $[;]$ stands for concatenation,  $\textbf{W}_0\in\mathbb{R}^{1 \times {h+1}}$ is a GELU activated fully connected layer with $h$ hidden dimensions and 
$\textbf{W}_1\in\mathbb{R}^{h+1 \times 3}$ is the final classification layer activated with Softmax for 3 classes.  
Equation \ref{eqn:model} represents the case where the network only has a single hidden layer $l$.  

We develop three different versions of \myClassifier. The first handles short texts (title articles), the second long texts (full articles), and the third also leverages domain credibility scores.

For ``short text'' experiments, we explore pre-trained Transformer-based language models such as T5 \cite{raffel2020exploring-t5}, DeBERTa \cite{he2020deberta} and CLIP \cite{radford2021learning-clip}. 
For our experiments on full article texts, we employ transformer-based models tailored for processing larger text sequences.
More specifically, we utilize Longformer \cite{beltagy2020longformer}, a model that integrates both local (window-based) and global attention mechanisms. 
For feature extraction, we utilize the [CLS] token to capture contextual information and train our classifier.
Finally, we similarly employ LongT5 \cite{guo2021longt5}, an extension of the T5 model, suitable for long texts.

In order to incorporate DCS into \myClassifier, we encode Factuality Scores categories `satire'(-3), `very low'(-2), `low'(-1), `mostly factual'(2), `high'(3) and `very high'(4). Articles without factuality scores are assigned a value of 0. For articles with a `mixed' score, we consider the MBFC Credibility Rating. If it indicates `medium credibility', `mixed' is mapped to 1. Conversely, if it is `high credibility' or `low credibility', `mixed' is mapped to 2 or -1, respectively. 
This differentiation ensures that we account for each case of `mixed' credibility and the absence of data. 
Finally, we normalize DCS into a range of (0,1), concatenate them with the text embeddings and pass the combined input through \myClassifier.

\subsection{Implementation Details}

We implement \myClassifier\ using the PyTorch deep learning framework, leveraging its efficiency for training neural networks. To ensure reproducibility, we set the random seed to 42 before conducting any experiments. The dataset is divided into three subsets using an 80/10/10 training/validation/test split.

For both short- and long-text models, we utilize 3-fold Cross Validation and Grid Search, respectively to optimize and fine-tune \myClassifier. 
In both cases, we explore various configurations including hidden sizes $h \in \{{512,1024\}}$ and number of layers $l \in \{1, 2, 3\}$. We employ the Adam optimizer with learning rates $lr \in \{1e-3, 5e-4, 1e-4, 5e-5\}$ and batch sizes $b \in \{512, 1024, 2048 \}$. 
Additionally, we experiment with dropout rates $d \in \{0.1, 0.2, 0.25\}$
and L2 regularization $r \in \{0, 1e-2, 1e-3\}$
to prevent overfitting and improve generalization.

Finally, we experiment with baseline models, namely Logistic Regression, Naive Bayes, Decision Trees and Multi-layer Perceptron (MLP) trained only on domain credibility scores or on statistical feature extraction methods like CountVectorizer and TF-IDF. 

\section{Results}

\subsection{Quantitative Results}

Table \ref{tab:baseline-experiments} demonstrates the results of the baseline classifiers. We observe that Count Vectorizer achieves 66.3\% accuracy with Logistic Regression and 69.4\% with the MLP Classifier while TF-IDF yields 67.0\% accuracy with Naive Bayes and 69.5\% with an MLP classifier.
When only leveraging 
DCS, a Decision Tree classifier reaches 68.5\% accuracy, closely competing with text-based models.

Table \ref{tab:experiment-results} illustrates the performance of \myClassifier\ while leveraging different transformer-based encoders.
We observe that with short texts (titles) and without DCS, \myClassifier\ reaches the best performance of 79.5\% accuracy while employing embeddings from DeBERTa\footnote{\url{https://huggingface.co/microsoft/deberta-base}} 
and having hidden dimensions $h=[512,1024,1024]$, learning rate $lr = 5e-5$, dropout $d=0.1$, l2 regulation of $r=1e-3$ and batch size $b=1024$. 
This performance is followed by integrating CLIP's text encoder \footnote{\url{https://huggingface.co/openai/clip-vit-base-patch32}} 
reaching 79.3\% accuracy and then T5 \footnote{\url{https://huggingface.co/google-t5/t5-large}} reaching 78.3\% accuracy.
When employing long texts (full articles), \myClassifier\ with 
Longformer \footnote{\url{https://huggingface.co/allenai/longformer-base-4096}} as the backbone encoder,
achieves 89\% accuracy with $h = [1024,1024,1024]$, $lr = 5e-4$,  $d=0.2$, $r=1e-3$ and $b=2048$. 

\begin{table}
\centering
  \caption{Baseline Experiments on \myDataset.}
  \label{tab:baseline-experiments}
  \begin{tabular}{ccc}
    \toprule
    Input&Classifier&Accuracy\\
    \midrule
     Count Vectorizer & Logistic Regression & 66.3\% \\
     Count Vectorizer & MLP Classifier & 69.4\% \\
     TF-IDF & Naive Bayes & 67.0\% \\
     TF-IDF & MLP Classifier & 69.5\% \\
     Domain Scores Only & Decision Trees & 68.5\% \\
  \bottomrule
\end{tabular}
\end{table}

\begin{table}
\centering
    \caption{Performance of \myClassifier\ on \myDataset\ for short or long texts, with different backbone encoder and with or without Domain Credibility Scores (DCS). 
    }
  \label{tab:experiment-results} 
  \begin{tabular}{cccc}
    \toprule
    Encoder&Accuracy w/o DCS &Accuracy w/ DCS&Input\\
    \midrule
     T5 & 78.3\% & 88.6\% & short texts \\
     Clip Text & 79.3\% & 91.0\% & short texts \\
     DeBERTa & 
     79.5\% & 91.5\% & short texts\\
     \midrule
     Long T5 & 79.1\% & 89.5\% & long texts \\
     Longformer & 89.0\% & 94.4\% & long texts \\
  \bottomrule
\end{tabular}
\end{table}

Furthermore, incorporating DCS from MBFC, 
can significantly and consistently enhance the classification accuracy of \myClassifier\ across all 5 Transformer encoders.
For short texts, the accuracy of \myClassifier\ with DeBERTa increases to 91.5\%, reflecting a notable relative improvement of +12\% while for long texts, Longformer facilitates a substantial boost, reaching 94.4\%, a +5.4\% relative improvement. Overall, these results demonstrate notable improvements in classification accuracy across all backbone model encoder, highlighting the effectiveness of incorporating domain-specific characteristics into \myClassifier.

\begin{table*}
\centering
\footnotesize
  \caption{Inference Results: Applying EVVER-Net on evidence from various datasets.}
  \label{tab:Inference-Results}
  \begin{tabular}{cccccc}
    \toprule
    Dataset & Data Type & Fact-checked & Credible & Unreliable & Samples\\
    \midrule
     LIAR-PLUS&
     Ruling statements of fact-checked articles
     &98.5\% &1.0\%&0.5\%&10,238\\
     
     MOCHEG&
     Evidence snippets from fact-checked articles
     &83.6\%&2.4\%&14.0\%&27,528\\
     
     FACTIFY&
     Article from ``Support'' \& ``Insufficient'' classes
     &6.4\%&88.0\%&5.6\%&34,000\\
     
     FACTIFY&
     Articles from ``Refute'' class 
     &95.0\%&3.5\%&1.5\%&8,500\\
     
     NewsCLIPpings+&
     Article titles (Web) 
     &14.1\%&64.5\%&21.4\%&45,907\\
     
     VERITE &
     Article titles (Web) - across 3 classes
     &35.3\%&42.6\%&22.1\%&1,611\\

     VERITE &
     Article titles (Web) - `True' and `Miscaptioned' classes
     &45.2\%&35.2\%&19.6\%&1,094\\

     VERITE &
     Article titles (Web) - `Out-of-Context' class
     &14.5\%&58.2\%&27.3\%&517\\
     
  \bottomrule
\end{tabular}
\end{table*}

\begin{figure*}
    \includegraphics[width=1\textwidth]{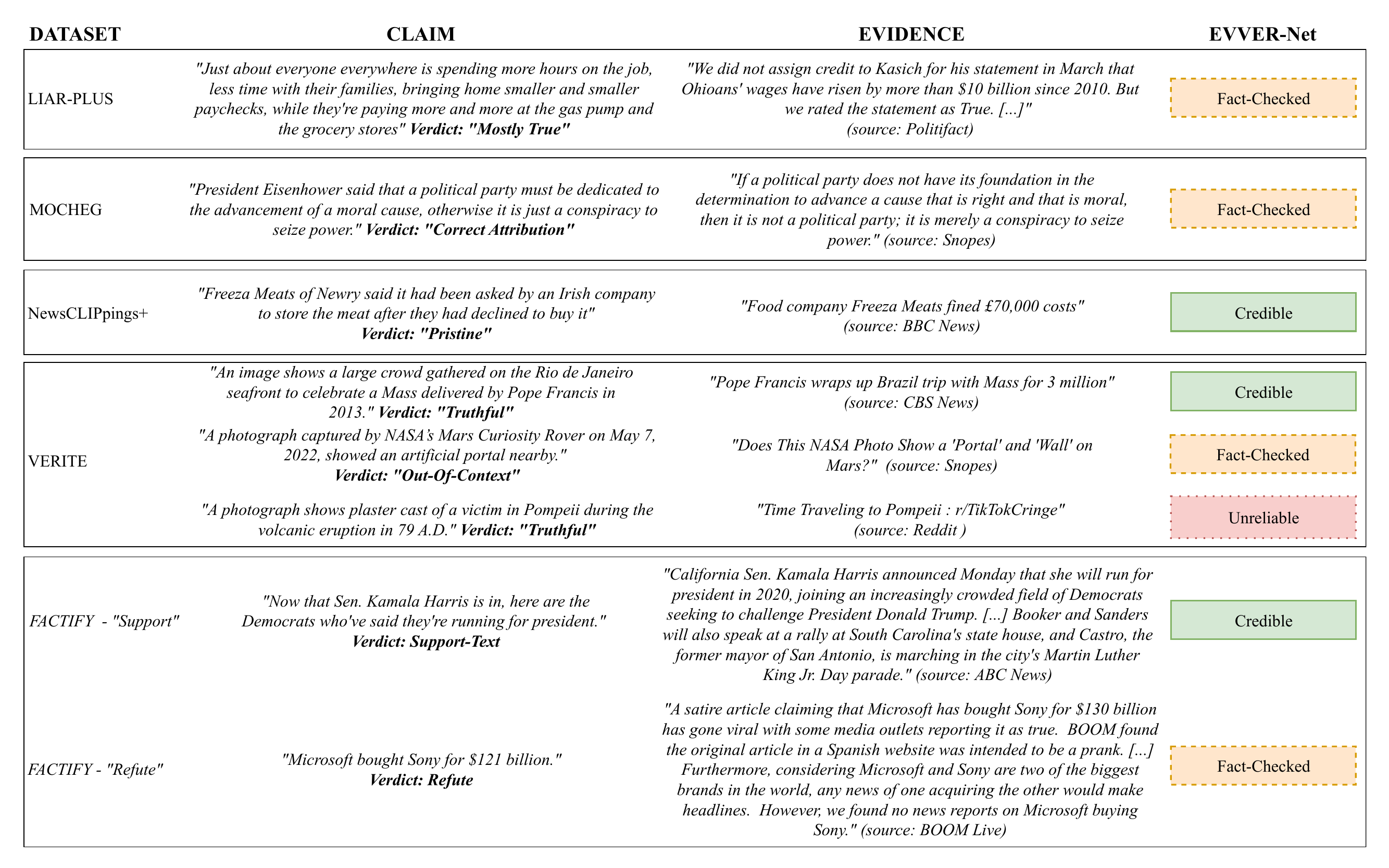}
    \caption{Inference examples from EVVER-Net applied on the evidence of various datasets. We manually included the domain names from which the evidence originates, as they are not provided by the datasets.}
    \label{fig:inference}
\end{figure*}

\subsection{Qualitative Analysis and Inference}

In this section, we apply \myClassifier\ to existing datasets, encompassing various text, multimodal, and Web-sourced datasets. 
We do not use DCS in this section because the domain names from which the evidence was collected are not provided by these datasets. 
By examining the classifier's performance across diverse datasets, we aim to assess its robustness and applicability in different contexts. Specifically, we focus on datasets containing fact-checked articles, as well as those incorporating external evidence from various sources on the Web.
Our evaluation begins with testing on text datasets, where the classifier's ability to discern credible information from unreliable is put to the test. We then extend our analysis to multimodal datasets and finally, we explore datasets that aggregate evidence from the Web, mirroring the real-world application scenario of \myClassifier. 
Results are summarized in Table \ref{tab:Inference-Results} 

\subsubsection{LIAR-PLUS Dataset}
We initiate our evaluation with the LIAR-PLUS dataset \cite{alhindi-etal-2018-evidence}, a well-known repository of fact-checked articles. 
LIAR-PLUS provides not only labeled articles but also detailed justifications for the veracity of each claim, which are sentences extracted from the section `Our Ruling' of each fact-checked article. 
Therefore, we would expect high rates of ``fact-checked'' class. 
Indeed, when we extract features with DeBERTa and apply \myClassifier\ on the ``justification evidence'', it successfully recognizes fact-checked evidence pieces, with 98.5\% being classified as Fact-checked, 1\% as credible, and 0.5\% as Unreliable. 
These findings underscore \myClassifier\ ability to discern information from fact-check websites, affirming its utility in real-world fact-checking applications.

\subsubsection{MOCHEG Dataset} 
We further extend our evaluation by applying \myClassifier\ to the MOCHEG dataset \cite{mocheg}, which aggregates data from Politifact and Snopes websites. 
Specifically, we focus on the `evidence' column of Corpus 2, comprising highlighted text snippets from fact-checked articles. 
Again, 
we observe consistent performance, with rate of 83.6\% of evidence pieces being classified as fact-checks.

\subsubsection{FACTIFY Dataset} 
The next dataset we evaluate is the FACTIFY dataset \cite{mishra2022factify}. 
This comprises textual and image claims, each associated with a reliable source of information referred to as a `document'. Claims are categorized into three classes: support, insufficient, and refute, based on their relationship with the corresponding document. 
Authors collected and scraped news articles from various credible sources in the US and India to compile data for the support and insufficient categories. We extract features with Longformer from the full-text article bodies and pass them through our classifier. 
Articles from the ``Support'' and ``Insufficient'' classes were collected from credible sources while articles in the ``Refute'' classes were collected from fact-checking websites. 
Our analysis reveal that 88\% of the contents categorized as ``Support'' or ``Insufficient'' are classified as credible by the classifier, while only 6.4\% and 5.6\% are classified as Fact-checked and Unreliable, respectively. 
Notably, \myClassifier\ correctly identifies that 95\% of the documents in the ``Refute'' class as Fact-checked.

\subsubsection{NewsCLIPpings+ Dataset} 
We extend our evaluation to include the evidence sourced from \cite{abdelnabi2022open}, referred to as NewsCLIPpings+, which consists of external information collected from the Web. 
Abdelnabi et al. \cite{abdelnabi2022open} utilized the NewsCLIPpings dataset \cite{luo2021newsclippings}, which contains both pristine and falsified (algorithmically decontextualized) image-caption pairs.
In this setup, textual evidence was obtained by querying images in an inverse search mode using the Google Vision API. 
For our analysis, we utilize the test set of NewsCLIPpings+, comprising 7,264 image captions and 51,799 scraped text evidence pieces. After removing inaccurately scraped entries, such as image file names or `Page Not Found' entries, we are left with 45,907 textual evidence pieces.
\myClassifier\ estimates that 64.5\% of the evidence is Credible, 21.4\% Unreliable, and 14.1\% Fact-checked. 
Despite the majority of the evidence being considered credible, the presence of some pieces sourced from unreliable or leaked origins in NewsCLIPpings+ may not only compromise the accuracy of the veracity predictions but also introduce unrealistic elements into the assessment process.

\subsubsection{VERITE Dataset}
We also assess evidence extracted in \cite{papadopoulos2023red} for the VERITE dataset \cite{papadopoulos2024verite}, which integrates image-caption pairs similar to the previous dataset. In this dataset, `MisCaptioned' pairs sourced from fact-checked articles like Snopes and Reuters were considered. The authors collected misleading claims along with their associated images, utilizing the Google API to retrieve textual and visual evidence. 
After cleaning the gathered evidence, we obtain 1,611 text snippets, gathered from querying  1,000 image captions, classified as `True', `Miscaptioned', or `Out-of-Context'. Our classifier categorizes these snippets with 42.6\% labeled as Credible, 35.3\% as Fact-checked, and 22.1\% as Unreliable.
It is important to note that the higher percentage of Fact-checked snippets in this dataset is due to sourcing claims and images directly from fact-checked articles in the `Miscaptioned' class. When these claims are searched using the Google API, it often returns the original or similar fact-checking articles, resulting in leaked evidence. For image-caption pairs classified as `Out-of-Context', we observe a lower Fact-checked (14.5\%) and a higher Unreliable percentage (27.3\%) compared to the other categories. In this case, the claims are taken from fact-checked articles but the out-of-context images are retrieved from the Web. Thereafter, these images are used to retrieve textual information from a Google API thus increasing the likelihood of unreliable evidence.

\subsubsection{Inference}
Our findings reveal that widely-used datasets incorporate leaked and unreliable evidence during the AFC process. 
This highlights the critical need for robust filtering mechanisms, like \myClassifier, to identify and exclude such information effectively. 
Figure \ref{fig:inference} provides examples sourced from LIAR-PLUS, MOCHEG, NewsCLIPpings+, VERITE, and FACTIFY datasets, alongside their classification by \myClassifier.
For instance, consider how ``[...] But we rated the statement as True.'' (LIAR-PLUS) or ``A satire article [...] intended to be a prank [...]'' (FACTIFY) - taken directly from within fact-checked articles - provide information curated by fact-checkers that directly support or refute the claim, thus creating an unrealistic scenario for early detection of new misinformation. 
On the other hand, short article titles such as ``Does This NASA Photo Show a `Portal' and `Wall' on Mars?'' or ``Time Traveling to Pompeii: r/TikTokCringe'' (VERITE), does not necessarily provide any unreliable or leaked information by themselves. 
However, if the articles' content were to be collected and analysed, they would be problematic for AFC systems.

\section{Conclusion}

In this study, we address the critical but overlooked challenges associated with verifying the quality of external information used as evidence in existing datasets, particularly the presence of leaked and unreliable information. We develop the \myDataset\ dataset, comprising 91,632 news articles classified as Fact-checked, Credible, and Unreliable. Additionally, we introduce the Evidence Verification Network (\myClassifier), a robust solution for evidence verification and filtering, and conducted experiments with various language models.
\myClassifier\ reaches 79.5\% and 89.0\% accuracy for short and long texts, respectively, without utilizing domain credibility scores. 
By utilizing domain credibility scores, the performance of \myClassifier\ further improves to 91.5\% and 94.4\%.
Furthermore, our analysis on widely used datasets, including LIAR-PLUS, MOCHEG, FACTIFY, NewsCLIPpings+ and VERITE, reveals that collected evidence often contain leaked and unreliable information, thereby diminishing the effectiveness and realism of building robust AFC systems. These findings underscore the importance of implementing evidence verification and filtering solutions such as EVVER-Net during the evidence retrieval task of AFC systems.

While our study offers valuable insights into evidence verification in AFC, it has certain limitations. First, it solely focuses on textual evidence, overlooking the potential influence of images and videos. Visual elements
can also contain unreliable information (e.g., by being fabricated, manipulated or synthetically generated) or having artifacts that indirectly indicate that they are sourced from fact-checked articles (e.g., watermarks). 
Future research should aim to enhance the robustness of evidence verification and filtering methods like EVVER-Net by expanding the CREDULE dataset to include a more diverse range of articles and sources, including multimedia content. 
Furthermore, we refrain from collecting new external information from the Web and filtering them with \myClassifier\  in order to examine its impact on AFC systems. 
We hypothesize that the performance of AFC systems would decrease if they previously primarily relied on leaked evidence, which would afterwards be removed by \myClassifier. 
On the other hand, we hypothesize that removing unreliable information would improve performance, thus somewhat balancing out the decrease in performance from removing leaked evidence.  
Overall, this would result into a more realistic and reliable framework for training and evaluating AFC systems, especially on new and emerging misinformation. 
Nevertheless, future research should systematically examine these effects. 

\section*{Acknowledgements}
This work is partially funded by the Horizon Europe projects vera.ai under grant agreement no. 101070093 and DisAI under grant agreement no. 101079164.

\bibliographystyle{unsrt}  
\bibliography{references}

\end{document}